\pdfoutput=1
\documentclass[11pt]{article}

\usepackage{ACL2023}
\usepackage{tikz-dependency}

\usepackage{times}
\usepackage{latexsym}

\usepackage[T1]{fontenc}
\newcommand{\yg}[1]{\textcolor{purple}{YG: #1}}

\usepackage[utf8]{inputenc}

\usepackage{microtype}

\usepackage{inconsolata}

\usepackage{color,soul}
\usepackage{graphics}

\title{Two Kinds of Recall}

 \author{Yoav Goldberg \\
         Bar-Ilan University \and Allen Institute for AI \\
         \texttt{yoav.goldberg@gmail.com} \\
         @yoavgo
}
\author{Yoav Goldberg \\
         Bar-Ilan University \and Allen Institute for AI \\
         \texttt{yoav.goldberg@gmail.com} \\
         twitter: \texttt{@yoavgo} \\ \\
         \textcolor{gray!70}{\textbf{This paper was rejected from ACL-2023. See reviews after the references.}}
}

\begin{document}
\maketitle
\begin{abstract}
It is an established assumption that pattern-based models are good at precision, while learning based models are better at recall. But is that really the case? I argue that there are two kinds of recall: d-recall, reflecting diversity, and e-recall, reflecting exhaustiveness. I demonstrate through experiments that while neural methods are indeed significantly better at d-recall, it is sometimes the case that pattern-based methods are still substantially better at e-recall. Ideal methods should aim for both kinds, and this ideal should in turn be reflected in our evaluations.
 
\end{abstract}

\section{Introduction}
A common folk wisdom in applied natural language processing (NLP) is that there is an inherent tradeoff between pattern-based methods and machine-learning based methods, in which the pattern-based methods are more precise, while the learning-based methods identify more cases (have better recall). With recent advances in neural-network based models, the learning based methods became significantly more precise, leaving them with both good precision \emph{and} better recall, compared to the pattern based methods.

However, I argue that this story is incomplete, and that pattern-based methods, \emph{while being significantly worse at recall} than learning-based methods, \emph{can also at the same time be significantly better at recall} than learning-based methods. This is possible due to the existence of two distinct kind of recall. The first kind of recall favors \emph{diversity}: finding many variations of the same main theme. Learning-based methods indeed excel in this kind of recall. The second kind of recall favors \emph{exhaustiveness}: once a variation is known, finding all occurrences of instances of this kind. Here, pattern-based methods shine, while learning-based methods sometimes still struggle, or---even at the best case---are less reliable than pattern-based methods.

Thus, I argue that when evaluating systems for recall, we should distinguish between the two kinds, \textbf{d-recall}, reflecting diversity, and \textbf{e-recall}, reflecting exhaustiveness. Ideal systems should achieve both kinds of recall (while remaining precise). However, current datasets and evaluation methods focus primarily on d-recall, while leaving e-recall as an incidental afterthought.

I will demonstrate the two kinds of recall and the tendencies of neural vs. pattern based systems on a task that combines sentence retrieval with relation extraction: identifying sentences that contain an instance of some binary relation that holds between two sentence entities.

The neural system is a RoBERTA-large\footnote{\cite{roberta}} based Question Answering model, trained on the SQuAD v.2.0 dataset.\footnote{\cite{rajpurkar-etal-2018-know}} The pattern based system will rely on pattern matching over syntactic dependency trees (which are automatically obtained through a neural-based parser).

Note, however, the the aim of this paper is not to compare neural and pattern-based methods for information retrieval, but rather to highlight the lack of systematicity of learning-based methods, which result in sub-optimal e-recall, and to shine a light on e-recall as an important property of language learning systems which is currently not being properly evaluated. While the case-study here is specific to information extraction, I believe the general trend holds (but may be harder to show) for all extractive NLP tasks, including question-answering, coreference resolution, named entity recognition, and others.

\section{Background}

\subsection{Dependency Trees and Syntactic Patterns}

Sentences have a syntactic structure that governs how the different words relate to each other through various constructions \cite{valin2001}. Syntactic theories define these structures, and syntactic parsers recover them from sentences. The relations between words are structural rather than semantic: phrasing that mean the same thing, such as ``Sam closed the door'', ``The door was closed by Sam'', ``Sam, who closed the door'' and ``The door which was closed by Sam'' all have different syntactic structures. 

In this piece I focus on \emph{dependency-based} syntactic analysis \cite{kubler2009}, in which each word is a node in a directed tree, where some nodes are parents of other nodes. The edges between tree nodes are labeled, reflecting the syntactic relation that holds between the words. These are the syntactic structures of two of the sentences above, according to version 1.0 of the \emph{universal dependencies} syntactic analysis scheme \cite{nivre-etal-2016-universal}.\\

\scalebox{1}{
        \begin{dependency}[theme = default]
            \begin{deptext}[column sep=0.2cm]
                Sam \& closed \& the \& door \\
            \end{deptext}
            \depedge[edge above,edge height=3ex]{2}{1}{nsubj}
            \depedge[edge above,edge height=3ex]{2}{4}{dobj}
            \depedge[edge above,edge height=1ex]{4}{3}{det}
            %\depedge[edge below]{6}{5}{det}{3ex}
        \end{dependency}
    }

\scalebox{1}{
        \begin{dependency}[theme = default]
            \begin{deptext}[column sep=0.2cm]
                The \& door \& was \& closed \& by \& Sam \\
            \end{deptext}
            \depedge[edge above,edge height=1ex]{2}{1}{det}
            \depedge[edge above,edge height=3ex]{4}{2}{nsubjpass}
            \depedge[edge above,edge height=1ex]{4}{3}{aux}
            \depedge[edge above,edge height=3ex]{4}{6}{nmod:by}{3ex}
            \depedge[edge above,edge height=1ex]{6}{5}{case}
            %\depedge[edge below]{6}{5}{det}{3ex}
        \end{dependency}
    }

%\scalebox{1}{
%        \begin{dependency}[theme = default]
%            \begin{deptext}[column sep=0.2cm]
%                Sam \& , \& who \& closed \& the \& door \\
%            \end{deptext}
%            \depedge[edge above,edge height=1ex]{4}{3}{nsubj}
%            \depedge[edge above,edge height=3ex]{4}{6}{dobj}
%            \depedge[edge above,edge height=1ex]{6}{5}{det}
%            \depedge[edge above,edge height=3ex]{1}{4}{acl:relcl}
%            \end{dependency}
%    }
    
%\scalebox{1}{
%        \begin{dependency}[theme = default]
%            \begin{deptext}[column sep=0.2cm]
%                The \& door \& which \& was \& closed \& by \& Sam \\
%            \end{deptext}
%            \depedge[edge above,edge height=1ex]{2}{1}{det}
%            \depedge[edge above,edge height=3ex]{5}{3}{nsubjpass}
%            \depedge[edge above,edge height=1ex]{5}{4}{aux}
%            \depedge[edge above,edge height=3ex]{5}{7}{nmod:by}{3ex}
%            \depedge[edge above,edge height=1ex]{7}{6}{case}
%            \depedge[edge above,edge height=5ex]{2}{5}{acl:relcl}
%          %\depedge[edge below]{6}{5}{det}{3ex}
%        \end{dependency}
%    }

Neural-network based automatic parsers are good at recovering the underlying syntactic structure of sentences, with high accuracy and high consistency \cite{kiperwasser-goldberg-2016-simple,dozat2017,spacy2}. 

\paragraph{Syntactic Patterns}

Some paths in a syntax tree correspond to semantic relations. Thus, one can extract information from text by first obtaining syntactic parse trees, and then looking for specific patterns in these trees. For example, the pattern:

\scalebox{1}{
        \begin{dependency}[theme=default]
            \begin{deptext}[column sep=0.2cm]
                \_\_\_$_{a1}$ \& was \& educated \& at \& \_\_\_$_{a2}$ \\
            \end{deptext}
            \depedge[edge above,edge height=3ex]{3}{1}{nsubjpass}
            \depedge[edge above,edge height=1ex]{3}{2}{aux}
            \depedge[edge above,edge height=3ex]{3}{5}{nmod:at}
            \depedge[edge above,edge height=1ex]{5}{4}{case}{3ex}
          %\depedge[edge below]{6}{5}{det}{3ex}
        \end{dependency}
    }

will match sentences whose syntactic trees contain the above subtree. This includes sentences such as:

\noindent 1. \emph{\underline{He}$_{a1}$ was born in London , and \textbf{was} \textbf{educated} \textbf{at} \underline{University College , London}$_{a2}$, of which he subsequently became a fellow .}

\noindent 2. \emph{Born to a Sikh family on 4 May 1922~, \underline{Talwar}$_{a1}$ \textbf{was} \textbf{educated} for a short time \textbf{at} \underline{the Lawrence} \underline{School , Sanawar}$_{a2}$ , along with his brother Rana Talwar .}

\noindent 3. \emph{\underline{Paul}$_{a1}$ \textbf{was} primarily \textbf{educated} \textbf{at} \underline{St. Jarlath 's} \underline{Vocational School}$_{a2}$ .}

We can further restrict the pattern, requiring for example for the subject position $a1$ (indicated by an incoming nsubjpass edge) to contain a proper-noun and not a pronoun, ruling out sentence (1) above.

Syntactic patterns are interpretable, efficient and precise, but have a clear drawback: there are many ways to express the same semantic relation, and we need to tailor a distinct pattern for each one. In other words, \emph{the patterns have low d-recall}.

\subsection{Extractive Question Answering and the SQuAD dataset}

SQuAD (``Stanford Question Answering Dataset'') \cite{rajpurkar-etal-2016-squad,rajpurkar-etal-2018-know} is a collection of over 150,000 <question, passage> pairs, where the answer to the question is a span in passage. The SQuAD dataset is used to train machine learning models to perform \emph{extractive QA}: given a question and a passage, mark the passage span that answers the question. In the real world, not every passage contains an answer to every question. To make the task more realistic, the SQuAD v 2.0 dataset \cite{rajpurkar-etal-2018-know} also includes <question, passage> pairs where the question is not answerable by the passage. Now, the model is tasked with identifying the answer span, or indicating that no relevant answer span exists (``no answer''). The answerable question-passage pairs were created by showing human annotators a passage, and asking them to ask a question and mark its answer in the text. The annotators were also encouraged to not repeat the passage phrasing in their questions. 

The result is a set of many entity-centric questions, where the question contains an entity and the answer span contains a related entity. For example,  ``Which gas makes up 20.8\% of the Earth's atmosphere?'' (answer: diatomic oxygen). The SQuAD dataset and the extractive-QA task became very popular, and models ``solve'' the SQuAD benchmark with very high %(above human) 
accuracies.\footnote{See e.g. systems' scores at \url{https://rajpurkar.github.io/SQuAD-explorer/})} %(e.g. \cite{squad-sota1} as well as other models at the SQuAD leaderboard at \url{https://rajpurkar.github.io/SQuAD-explorer/}).
%\footnote{How can a text understanding model achieve above-human accuracy? One answer is that the test measures something beyond text understanding. For example, sometimes various spans can answer the question and constitute valid answers (e.g., ``diatomic oxygen'', ``diatomic oxygen gas'' and ``the diatomic oxygen gas'' are all valid answers to ``which fas makes up 20.8\% of ...''). A model can learn to identify that tendencies of span boundary selection, while humans will be much less consistent, or will be consistent in a way which differs from the dataset, and be penalized from it. Nonetheless, regardless of these small evaluation issues, SQuAD based QA models perform remarkably well on the benchmark.} 
As a result, SQuaD-based models are often used as of-the-shelf NLP components. For example, the model \texttt{deepset/roberta-base-squad2} available on Huggingface models hub,\footnote{\url{https://huggingface.co/models}} has been downloaded over 650,000 times in Jan 2023 alone, and the more accurate but also more compute intensive \texttt{deepset/roberta-large-squad2} was downloaded over 35,000 times in the same month.

\section{Experiments}

The following experiments are all performed through the SPIKE system \cite{shlain-etal-2020-syntactic,taub-tabib-etal-2020-interactive}, available at \url{https://spike.apps.allenai.org/}. The system allows to perform various query types over corpora which is pre-annotated with linguistic structures (including syntax, POS-tags, lemmas and named entities). The syntactic annotation in the SPIKE-indexed corpora are based on the universal dependencies v1 scheme, enhanced with additional (non-tree) arcs by the pattern-based pyBART system \cite{tiktinsky-etal-2020-pybart}. The SQuAD experiments are based on a model downloaded from Huggingface model hub. The model is \texttt{deepset/roberta-large-squad2}, the largest version of the most-popular squad2 model on huggingface. This model is based on the RoBERTA large pre-trained transformer \cite{roberta}, and is considered to be a strong extractive question-answering model.

\subsection{Demonstrating d-Recall}

In the first experiment, I retrieve a collection of PubMed sentences containing the words ``pain'' and ``molecule''. I treat each of the 380 returned sentences as a passage in a QA setup, and ask the SQuAD model ``which molecule relates to pain?''. This leaves us with 282 sentences marked as No-Answer, and additional 98 sentences in which the QA model identified an answer span. Most of these answer spans are correct, and the answers are diverse in terms of their underlying structures and words, for example:\\
\vspace{-0.7em}

\noindent 1. \emph{... we present evidence that ... \textbf{molecule} , \underline{nitric oxide} ( NO ) may act at several levels of the nervous system during the development of experimental neuropathic \textbf{pain} .}

\noindent 2. \emph{\underline{TRPA1} , a member of the transient receptor potential channel ( TRP ) family , is genetically linked to \textbf{pain} in humans , and ... }

\noindent 3. \emph{\underline{TRPV1} : an important molecule involved in the peripheral sensitization during chronic \textbf{pain} .}

\noindent 4. \emph{The \underline{TRPV1} is a well-known \textbf{pain} transducer \textbf{molecule} and plays crucial roles in the perception of inflammatory and thermal \textbf{pain} .}\\
\vspace{-0.7em}

In a similar experiment, I retrieve Wikipedia sentences including the words ``police'', ``arrest'', and a named person, and ask the QA engine ``who escaped?''. This results in answers such as:\\
\vspace{-0.7em}

\noindent 1. \emph{... reports Debbie to \textbf{police} , and \underline{Debbie} is last seen running down an alley to escape after officers arrive at the Gallagher house to \textbf{arrest} her for ...}

\noindent 2. \emph{She calls the \textbf{police} with a story about how \underline{Dirk} shot Johnny and then ran away , but they \textbf{arrest} her when they ...}\\[0.1em]
\vspace{-0.7em}

The diversity of these answers clearly demonstrate the d-recall abilities of the neural QA models. Extracting these answers using syntactic patterns would be a non-trivial challenge.

\subsection{Demonstrating (lack of) e-Recall}

In the second set of experiments, I start with queries which are based on syntactic patterns.
For the first query, I queried for the syntactic pattern:

\_\_\_\_$_{a1}$\emph{ was educated at }\_\_\_\_$_{a2}$ \\ demonstrated above, restricting the $a1$ position to be a person's name.

Running this query over the Wikipedia corpus results in a huge number of results, of which we take the first 5,000 matches. By design, all the results clearly indicate the ``PERSON educated at PLACE/School'' relation, via a recurring syntactic pattern. We now ask the SQuAD model, on each of these sentences,

\begin{figure*}[t!]
\includegraphics[width=\textwidth]{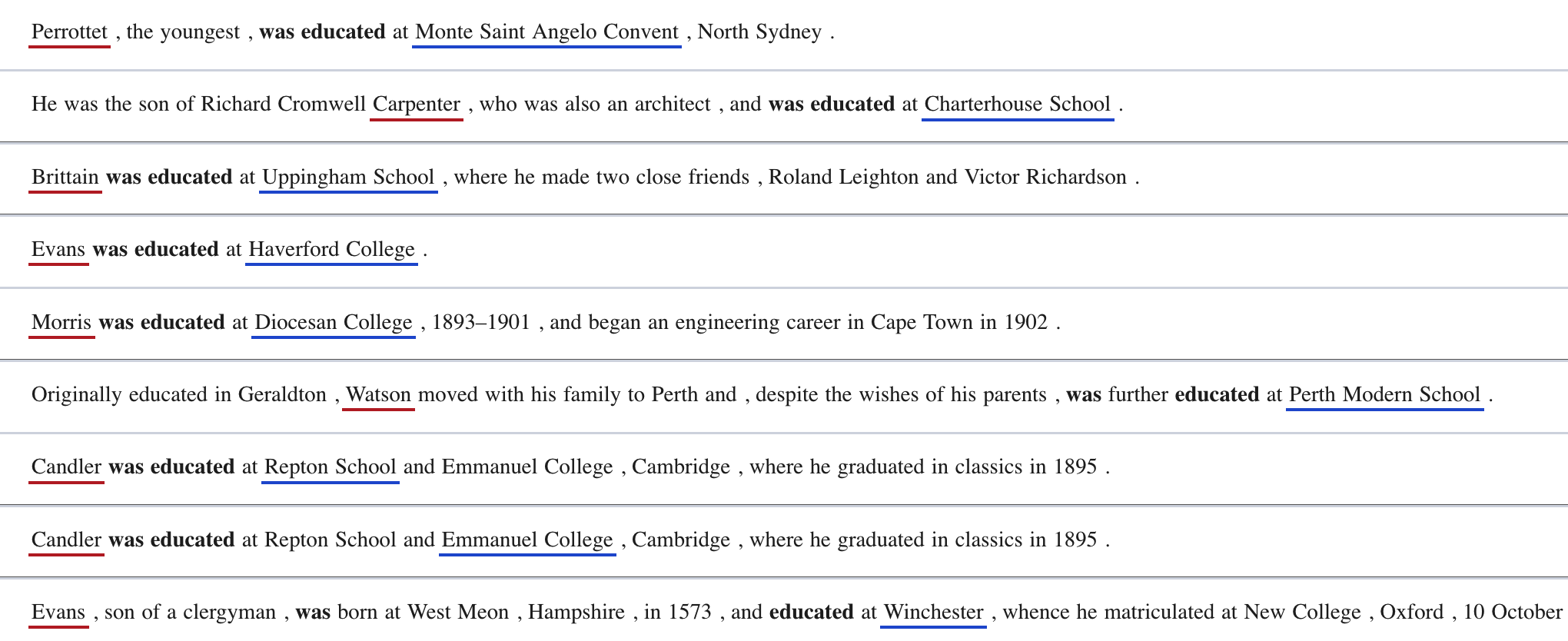}
\caption{Sample sentences on which the \textbf{roberta-large-squad2 model returned No-Answer for the question ``who was educated?''}. The marked spans are based on the syntactic-pattern match.}
\label{fig:educated}
\end{figure*}

``Who was educated?''

\noindent While the SQuAD model got many answers correctly, \textbf{it also identified 611 of the sentences as No-Answer cases}. Inspecting these sentences reveal that they all contain explicit and clear answers to the question. A sample of such No-Answer sentences is given in Figure \ref{fig:educated}. Note that the set on which the SQuAD model returns \emph{No-Answer} includes very direct and concise sentences, such as:\\
\vspace{-0.7em}

- \emph{Evans was educated at Haverford College}.\\
\vspace{-0.7em}

Question variations such as ``Who was educated somewhere?'', ``Who was educated at a school?'', ``Who was educated at a university?'', ``Who studied at a university?'' etc, result in very similar trends, with hundreds to thousands (!!) of No-Answer results for each question.\footnote{Additionally, in some of these question the model becomes less precise, marking spans such as ``Eton College'' and ``University of Glasgow'' as answers.}

\noindent Similarly, I ran the syntactic pattern:

\emph{\underline{DISEASE} was treated with \underline{CHEMICAL}}

\noindent over the PubMed corpus (DISEASE and CHEMICAL are entity-type restrictions), resulting in \textbf{3654} medical-condition / chemical pairs. Asking the QA model ``which condition was treated?'' results in \textbf{721 No-Answer cases}, including from sentences such as:\\
\vspace{-0.5em}

\noindent - \emph{\underline{The infection} \textbf{was} successfully \textbf{treated with} \underline{local voriconazole} followed by oral terbinafine}; and\\
- \emph{\underline{Early muscle pain} \textbf{was treated with} \underline{Ibuprofen}}.\\
\vspace{-0.5em}

Question variations such as ``what was treated?'' result in a similar number of No-Answer cases, where the the sets of No-Answer cases overlap, but are not the same.

This trend by which the QA model fails to identify many answers recovered by the syntactic pattern query is remarkably consistent for pretty much all syntactic patterns I tried.

There are possible explanations for the SQuAD model behavior: for example, the questions refer to more abstract concepts than what the model is used to, and the model's training data led it to be especially careful around potential answers where the text around the answer span is too similar to the question text. 

Yet, regardless of these reasons, the observed behavior remains: the model manages to correctly identify the answer is some of the cases, but also fails to identify the answer on very many others: it is failing in e-recall. The pattern based methods, in contrast, are naturally exhaustive in these cases, and can consistently identify cases which are not identified by the neural QA model.

\subsection{...but did you try GPT-3?}
When working on the final version of this paper, on March 18, 2023, I also performed a quick test with the \texttt{text-davinci-003} model. For the first 100 wikipedia results for the \emph{PERSON was educated at} pattern, I queries the model with the following prompt:

\noindent\texttt{based on the following text, who was educated somewhere? write only the name, and "None" if no one was educated.\\ text: \{text\}}

While \texttt{text-davinci-003} is obviously a very strong model, it still fails to demonstrate perfect e-recall. Out of the 100 sentences with a very clear person-educated-at pattern, \texttt{text-davinci-003} \textit{failed to extract} the person name from 2 of them, extracting the school name instead. 

\noindent\includegraphics[width=0.5\textwidth]{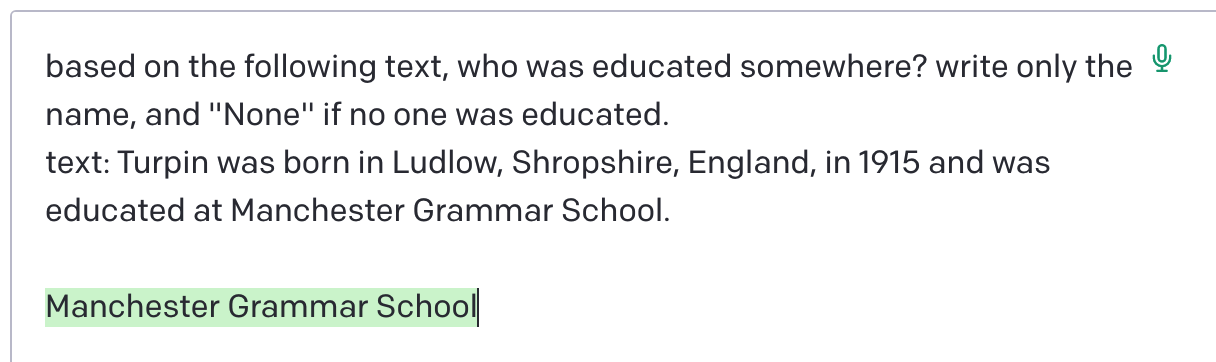}
\includegraphics[width=0.5\textwidth]{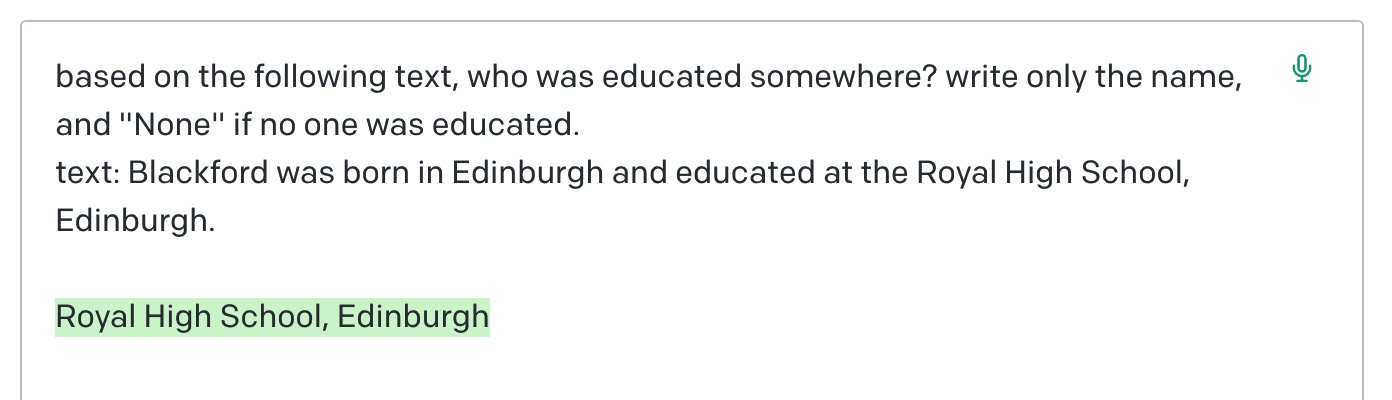}
%Turpin was born in Ludlow, Shropshire, England, in 1915 and was educated at \underline{Manchester Grammar School}.

While 98\% e-recall is impressive, we should note that it was over a set in which a pattern-based system obtained 100\% e-recall, and with a weird failure mode of identifying the wrong argument with a different semantic type.

\section{Discussion}
An obvious response to the SQuAD experiment above is ``we just need to add some results of the pattern-based queries to the QA model's training data''. Indeed, this will likely ``solve'' the issue described above, in the sense that it will be hard to demonstrate low e-recall of the resulting QA model, using the same technique. But I argue that such a ``solution'' will only hide the underlying problem, and not solve it. The above experiment demonstrated one class of instances in which the QA model is missing many valid answers. But do we know it is the only one? I argue that it is very likely \emph{not} the only e-recall failure mode of neural QA models, it is only one which was convenient to demonstrate. If we block this particular demonstration technique, it does not mean the models will stop having other blind-spots. 

%This observation is consistent with previous reports, in which neural models can learn to replicate a phenomena very well when measured on a test-set, but also fails in many subtle and hard to identify ways. For eaxmple, @@@gail@@@ and @@@kyle@@@. The e-recall failure demonstrated here another such case.

\section{Going Forward}
I have shown that there are two kinds of recall, one based on diversity, and another that is based on exhaustiveness. Currently, pattern-based methods still excel at ensuring exhaustiveness, while neural methods struggle. Moreover, while it is almost trivial to bound the exhaustiveness of pattern-based methods (for example, syntax-based methods are as exhaustive as the accuracy of the underlying parse trees), it is currently impossible to do so for learning-based methods.  And crucially, our evaluation methods don't test for exhaustiveness (and for an understandable reason: it is hard to do well, compared to diversity-based tests).

Recall of an ideal system should be both diverse \emph{and} exhaustive. Recognizing the existence of e-recall and keeping it in our minds as we develop future methods, is a good first step towards achieving this ideal. Vocabulary is important in establishing norms, and I would be happy if the e-recall and d-recall vocabulary (or a variation of it) would be adopted. However, while the recognition of a problem and the establishment of vocabulary around it is a good and necessary start, it is not sufficient. It is very hard to make progress on what we cannot reliably measure. This is thus a call for a new challenge to the community: developing robust methods of assessing e-recall, for a wide set of language understanding problems.
 
\section*{Limitations}

The work presents an opinion, and like any opinion, should be assessed and not taken for granted just because it is written. While the  opinion is backed by evidence, this evidence is reported as anecdotal cases, and not as a thorough quantitative study (for example, I could have repeated the experiment in 3.2 with hundred different syntactic patterns, and counted the number of no-answer cases in each. I chose not to do that. I hope the readers believe me that the results are indeed consistent across many patterns). This could be considered either is a limitation, or, as I rather see it, as a stylistic choice which is fitting of opinion papers like this one, who aim to make a point through demonstration, but where the point holds regardless of the exact numbers that could be measured.

\section*{Ethics Statement}

To the best of my judgement, this work does not raise any significant ethical concerns.

\section*{Acknowledgements}
This project has received funding from the European Research Council (ERC) under the European Union's Horizon 2020 research and innovation programme, grant agreement No. 802774 (iEXTRACT).

\bibliography{anthology,custom}
\bibliographystyle{acl_natbib}

\appendix

\clearpage

\section{Reviews and (some) Responses}

The paper received very low scores in ACL 2023, and I decided to post it on arxiv rather than going through the author-response process. Here are the reviews, with some responses inline. The aim of my responses is not to ``defend myself'' but rather to highlight differences between me and ``the community'' (as reflected by the reviewers) regarding our beliefs of what are desired qualities of a position paper. I hope to convince future reviewers of position-like papers to take a stance closer to the one I present here than to the one the reviewers took. I also answer questions where I believe readers will have a similar question to the reviewer, and may benefit from a clear answer.

The crux of the issue is that I attempted to highlight a deficiency of neural models and tell a story about evaluation and a metric which we do not currently evaluate (e-recall).
To demonstrate the point, I used an experiment tha anecdotally demonstrated the point. Indeed, the reviewers ``got'' the point. But they also focused on the more ``expected story'' of a competition between models. As such, they were looking for a more ``rigorous'' setup for the comparison, that will ``better convince'' them that ``the results hold''. But for me the results are just an anecdote, and are not the point. I argue that we should be able to let go of our fixation on numbers and comparisons, and try to identify broader stories when they exist. Especially in position-type papers.
(Some of the blame is on me for not making this clearer. In the current version of the paper, I added the last paragraph of the introduction, which was not present in the submitted version, in an attempt to state the point of the paper more explicitly.)

\subsection{Review 1}
\paragraph{What is this paper about and what contributions does it make?}
This paper argues that in NLP tasks, we should consider "two kinds of recall”: "d-recall,” which emphasizes the diversity of cases correctly identified by some system, and "e-recall,” which emphasizes the exhaustiveness of cases correctly identified by some system. This distinction is motivated by a supposed assumption that pattern-based systems for NLP are better for optimizing precision than neural systems, and neural systems are better for optimizing recall. I like to think of this distinction as follows: d-recall pertains to how diverse are the correctly identified variations within some class, while e-recall pertains to the proportion of some variation within a class that are correctly classified.
The paper presents some anecdotal evidence that if we think of d-recall and e-recall separately, we see that neural systems are good at d-recall, but less so at e-recall. The target task is relation extraction on sentences retrieved from a corpus, which is done through two comparison methods: 1) using a dependency parser-based retrieval mechanism, and 2) using a neural extractive question answering model which is a RoBERTa-based language model fine-tuned on SQuAD. Several experiments are performed by first using the parser-based model to retrieve N examples of some relation, e.g., "X was educated at Y” from some corpus, then applying the neural QA model. Then the author examines the outputs of the neural QA model to investigate their diversity and exhaustiveness.

Specifically, the author finds that if we consider a set of sentences of diverse syntactic structure, the neural QA model can correctly extract a target relation from a breadth of them. Meanwhile, if we retrieve sentences of the same syntactic structure, the neural QA model misses many target relations in them. This provides some evidence of the strength of the neural QA model with respect to diversity, but the brittleness of it with respect to exhaustiveness.

While I think the distinction of these "two kinds of recall” is novel and important in the context of these fine-tuned language models, which are widely known to spuriously change their outputs with small, inconsequential changes to the inputs, this paper does not put in the effort to warrant publication. The paper attempts to make conclusions based on anecdotal evidence, where there are missed opportunities for proper empirical studies. With more effort toward clear definitions of concepts introduced, and comprehensive empirical studies and analysis, this could be a substantial contribution.

\yg{Clearly the reviewer and myself see the role of position-papers differently. The experiments were meant to demonstrate the issue and convince people that it exists and is interesting, not to quantify it. As the reviewer notes later, they understood the concepts and found them interesting. I don't see any benefit in such quantification, and, as a reader, would have found them boring to read. I rather see concrete cases over aggregate numbers.}

\paragraph{Reasons to accept}
The paper proposes novel (to my knowledge) concepts of d-recall (recall favoring diversity) and e-recall (recall favoring exhaustiveness). This refinement of the recall metric could be a helpful way to measure the spuriousness of neural text classifiers.
While the experiments are not conclusive or complete, I found the problem formulation to be an interesting way to expose these two kinds of recall.

\paragraph{Reasons to reject}
The most glaring issue with this paper is that the two kinds of recall are not clearly defined evaluation metrics. I can think of a number of ways we can attempt to do this, e.g., by measuring the lexical or syntactic diversity of correctly classified examples within each class, but this effort was missing.

Almost all of the results presented are qualitative. The author only presents anecdotal evidence and cherry-picked examples, which is not conclusive and not rigorous enough for scientific publication. Even the quantitative results presented are not conclusive, as they are simply counts of supposedly misclassified examples by the neural QA model; but the data is not human-annotated and has no ground truth labels to my understanding, which means it could be generally noisier than the examples presented in the paper. This paper would be much more convincing if applied to some existing, human-annotated benchmarks, and we could calculate quantitative metrics to compare the d-recall and e-recall on different subsets of the data.

\yg{While ``scientific rigor'' is important in technical contributions, we should be careful not to throw the baby with the bathwater when insisting on it for position-like papers. The concepts of ``d-recall'' and ``e-recall'' I present are broader than just the information-extraction task I demonstrate them on. My intent was to raise awareness to a general phenomena, and hopefully ignite research around its solution. By providing a clear definition and evaluation metric, I'd be severely limiting the readers into a pre-specified solution, rather than letting them find their own, which may be better than mine (and I'd also be restricting them to the context of information-extraction).\\
\indent The call for using ``an existing, human annotated benchmark'' is also very limiting in this case, as existing benchmarks are not well suited for measuring e-recall (as stated in the intro ``\textit{current datasets and evaluation methods
focus primarily on d-recall, while leaving e-recall
as an incidental afterthought}''). More broadly, reliance on established benchmarks is both a driving force and a restraining force. We should be careful not to restrain too much.}

The paper only considers one task, one type of neural model, and one type of pattern-based model. In order to demonstrate the general conclusions made, a more comprehensive study may be necessary. If not, then the paper should be more clear about exactly where the conclusions apply.
\yg{This seems to view the paper as a claim about the relative performance of different models. This misses the broader---and to me substantially more important---message: there is a kind of recall that we as a community do not properly evaluate, and which current systems often fail on.}

\paragraph{Questions for the Author(s)}
- Question A: In the experiment mentioned in L204, where are the sentences retrieved from? How many were retrieved? \\
\yg{Who cares? Does knowing it change the picture in any meaningful way? The context is showing a bunch of sentences over which a SQuAD model works well, in-line with the established assumption in the community that SQuAD models work well. What does it matter where the sentences came from, or what the larger group size was? (the answer is Wikipedia, which was removed for submission version to fit in 4 pages, and then added back in the current version).} 

- Question B: Applying a model trained on SQuAD to single sentences is a bit out-of-domain, as these models are trained on full passages. Could that influence the trend you're seeing? Did you try any other types of neural models? \\
\yg{If it does, does it change the large picture in any way? If a model often work but sometimes fail when applied to a sentence rather than a paragraph, when the sentence is sufficient to answer the query, what does it tell us about the robustness of this model? Moreover, I'd argue that until we (NLP community) have a good notion of what a ``domain'' is, we should be careful in using ``domain shift'' as an excuse for model failures.}

- Question C: Is the comparison between pattern-based and neural approaches fair? You seem to apply the neural model to only sentences retrieved by the pattern-based approach. What do your results say about pattern-based approaches, if anything? \\
\yg{The results demonstrate that there is a set of sentences for which the neural model has low e-recall. Note that the reviewer is again tempted to force a story about model comparison, and not about two kinds of recall.}

- Question D: Why not try the experiments you suggest in the beginning of Section 4 then run some out-of-domain evaluation setting? Then you can actually support the claims here.\\
\yg{Which claims, and how will it help the two kinds of recall story?}

- Question E: RoBERTa is a bit old at this point. Do you plan to apply this study to any newer models to understand the variation with model size, training data size, etc.? \\
\yg{This continues to seek the ``let's compare models'' story, rather than the ``two kinds of recall'' story.} 

\paragraph{Missing References}
While it's a common thought that precision and recall can sometimes be at odds when building models for NLP tasks, I'm not sure I've heard of this assumption that pattern-based methods would be better at precision, and machine learning methods better at recall. While I don't find it surprising, a broad statement like this about the field should be justified if it will motivate the entire work. Please support this with some relevant references. I wasn't able to find any to point to myself…
This paper could use a proper literature review. I think some parts may be borrowed from Section 2 here.

\paragraph{Typos, Grammar, Style, and Presentation Improvements}
In the abstract, please give some specific ideas about the task, models, and final conclusions considered in this work.\\
\yg{The abstract lists the conclusion very clearly: \emph{``Ideal methods should aim for both kinds [of recall], and this
ideal should in turn be reflected in our evaluations''}. The reviewer is again trying to fit a model-comparison story, which should be ``scoped''.}

Recall is a metric used in all kinds of scientific disciplines, and this notion of "two kinds of recall” may not apply to all of them, so it would help to scope down a bit from the beginning.
In Section 1, a figure visualizing the difference between d-recall and e-recall would be helpful, or at least some more details about how e-recall is defined.

L087: "autoamtic” -> "automatic”

L093: grammatical error; remove "a”?

Section 2 could use a bit more attention to organization. First, it may be more appropriate to call it a "Methods” section rather than "Background.” Some introductory sentences in each subsection, and some conclusive sentences about how the syntactic patterns, SQuAD dataset, etc., are actually applied in the work would help the reader follow this section. Section 2.2 is one long paragraph - can we break it down a little?

L159: "experiements” -> "experiments”; "perfromed” -> "performed”

L164: "annoatated” -> "annotated”

I found the first paragraph of Section 3 hard to follow. In separate paragraphs, please introduce the pattern-based system implementation, and the neural-based system implementation.

Remove "(!!)” in L250.

The results in Section 3 need to be presented completely differently. There are too many changes to list here, but it would be much easier to follow if results were presented in a table and some actual evaluation metrics were calculated - don't just cherry pick examples and list counts of misclassified examples. Reduce the amount of qualitative discussions about results, e.g., at L246, L278, L282, and replace with proper empirical studies where possible. This is hard to follow and completely inconclusive.

Somewhere before Section 3, the problem should be formally defined. Section 3 is a bit hard to follow because the problem is not clearly defined.

\yg{(I fixed the typos and ignored the other stylistic suggestions.)}

\noindent Soundness:	1
\\ Excitement (Short Paper):	1.5
\\ Reviewer Confidence:	4
\\ Recommendation for Best Paper Award:	No
\\ Reproducibility:	3
\\ Ethical Concerns:	No

\yg{I wouldn't necessarily judge the soundness of a position paper by the exhaustiveness of its experiments.}

\subsection{Review 2}
\paragraph{What is this paper about and what contributions does it make?}
This paper discusses the properties of recall-based evaluation, distinguishing the recall into a "d-Recall" reflecting on response diversity and an "e-Recall" reflecting on response exhaustiveness. The author argues that while recent neural models are very good at d-Recall, they lack on e-Recall even compared to pattern-based methods. However, it is desirable to have systems whose responses are both diverse and exhaustive, i.e. complete. They present an example to support this assumption by showing high diversity of answers of a neural extractive QA model (RoBERTa-Large), and its inferiority on the completeness of the answers when asked to answer questions about contexts retrieved by a pattern-based method based on syntactic trees matching. Finally, the author argues that current evaluation methods favour diversity in evaluation and, thus, that future evaluation benchmarks should aim to take both aspects of recall into consideration.

\paragraph{Reasons to accept}
(i) The paper is easy to follow, giving priority to clear, short statements with minimal turns from the main line of argumentation. 
\\(ii) The distinction of "diversity" from "exhaustiveness" in the evaluation of systems responses makes sense to me. I think its distinguishing in evaluation testbeds could allow the community to identify important differences in the functioning of NLP systems while motivating future work in clearing out blind spots of some existing methods.

\paragraph{Reasons to reject}
I understand that a more extensive evaluation might fall outside the scope of a position paper, but having trouble relating to the main assumptions, I still had to rely on the evidence given in the example evaluations. My first two points refer to the problems that I identify in these.\\
(i) The paper supports "high d-Recall of neural systems" assumption by presenting the example of neural QA behaviour in responding to the question "which molecule relates to pain?" over a PubMed corpus and showing answers to these questions. This evaluation shows the performance of QA system trained on SQuAD under quite a large domain shift in both the question and contexts. Further, it evaluates the diversity of the responses subjectively, without comparison to the pattern-based responses. Therefore, from this case, it is hard for me to relate to the claim that the diversity of QA answers "clearly demonstrates" the diversity quality of neural models (L216).\\
\yg{This reviewer again focuses on the comparison between neural and pattern based system. And also, like reviewer 1 is suddenly having trouble believing that ``neural methods are good at diversity''.} 
\\(ii) "Lack of e-Recall" claim is in the paper supported by a high number of "not-answerable" responses of QA system on pattern-retrieved responses to some specific patterns. I am happy to believe that the situation is similar with other patterns (mentioned in Limitations, L342), but this approach also does not give a measured e-Recall of QA model into any context. Would the conclusions be the same even if the QA responds "no answer" for, say, 200 or 20 examples? I can not reconstruct any general conclusions from such an evaluation.\\
\yg{This infatuation with numbers again, and with comparisons / looking for thresholds. I really fail to see why the exact numbers matter here, or why a number of 200 will tell a different story w.r.t the existence of an e-recall failure of the neural model.}\\
Even assuming that the pattern-based methods return all correct responses, to show the difference, the same experiment could be done the other way around, to get some context: assuming that the QA responses (excluding "not-answerable") are all valid and matching them using the pattern-based method.
\yg{No, this would be measuring d-recall of the pattern-based system (which we know is bad).}\\
(iii) The paper's overall objective seems to be to motivate follow-up work to create benchmark(s) that will reflect equally on both kinds of recall. However, after reading the paper and looking at the examples, I have no idea how I would proceed towards that goal. This raises doubts in me on the usefulness of this work in a broader scope, as currently framed.\\
\yg{This is what makes the challenge a hard one, and why I think it warrants a position paper in the ``reality check'' theme track.}

\paragraph{Questions for the Author(s)}
- Question A: On L317, you state that "our evaluation methods don't test for exhaustiveness", but I am having difficulty to see at this point how this is the case. Can you give examples of this assumed blind spot on examples of specific, existing NLP benchmarks?\\
\yg{Models have very good performance on existing benchmarks. And benchmarks are not collected with many examples that follow a shared and simple pattern.}

- Question B: Can you be more specific in proposing the evaluation technique that would consider both kinds of recall while still being applicable to the broad coverage of NLP tasks?\\
\yg{I think this should be defined on a per-task basis. And if I knew the answer, I'd be writing a resources-and-evaluation paper and not a position-paper in the theme track.}

\paragraph{Typos, Grammar, Style, and Presentation Improvements}
Typos (some of which could be uncovered by simple type checking tools):
L56: SqUAD -> SQuAD

L93: Correspond to a semantic ... -> Correspond to semantic

L98: Will match -> will match

L150: SQuaD based-models -> SQuAD-based models

\noindent Soundness:	2
\\Excitement (Short Paper):	2.5
\\Reviewer Confidence:	3
\\Recommendation for Best Paper Award:	No
\\Reproducibility:	4
\\Ethical Concerns:	No

\subsection{Review 3}
\paragraph{What is this paper about and what contributions does it make?}
The paper explores the performance of pattern-based and learning-based models in terms of diversity and exhaustiveness of recall, introducing the concepts of d-recall and e-recall. Through experiments, the study suggests that neural methods are better at d-recall, while pattern-based methods are still substantially better at e-recall. The paper argues that ideal methods should aim for both kinds of recall and proposes that evaluations should reflect this ideal.

\paragraph{Reasons to accept}
Introduces the concepts of d-recall and e-recall, which could potentially be useful for evaluating and optimizing machine learning models in certain contexts.

Proposes that evaluations should aim for both kinds of recall, which could lead to more nuanced and comprehensive evaluations of machine learning models.

The paper is easy to follow

\paragraph{Reasons to reject}
The paper seems to present a specific and limited contribution, focusing on the performance of pattern-based and learning-based models on SQuAD in terms of d-recall and e-recall. While this may be an interesting and relevant problem, it is not clear how significant or generalizable the findings are, and whether they justify a full paper.

The paper does not provide enough context or related work to understand the novelty or importance of the proposed concepts of d-recall and e-recall.

Experiments are too simple to support that neural methods are better at d-recall, while pattern-based methods are still substantially better at e-recall. This is examined simply by two example questions: "which molecule relates to pain?” (for d-recall) and "Who was educated?” (for e-recall). I also feel why the pattern-based method can do e-recall better in this case is because pattern for "was educated at" is too simple. If the authors test on more difficult problems, the conclustion may be different.

Overall, the paper lacks depth, and the contributions and implications of the study are not well-established.

\paragraph{Questions for the Author(s)}
If we lower the threshold of prediction, could learning-based models achieve a higher e-recall? Is it meaningful to evaluate the recall without considering precision?
\yg{I evaluate e-recall on cases where the answer exists and is easily identified by a parser-based method. Changing the threshold will indeed result in fewer ``no answer'' cases, but will also be abysmall in precision on a more diverse dataset.}

\paragraph{Typos, Grammar, Style, and Presentation Improvements}
Line 222: a queries

\noindent Soundness:	2
\\ Excitement (Short Paper):	2
\\ Reviewer Confidence:	4
\\ Recommendation for Best Paper Award:	No
\\ Reproducibility:	4
\\ Ethical Concerns:	No

%\section{Example Appendix}
%\label{sec:appendix}

%This is a section in the appendix.

\end{document}